\documentclass[twocolumn]{article} % DO NOT CHANGE THIS
\usepackage{aaai25}  % DO NOT CHANGE THIS
\usepackage{times}  % DO NOT CHANGE THIS
\usepackage{helvet}  % DO NOT CHANGE THIS
\usepackage{courier}  % DO NOT CHANGE THIS
\usepackage[hyphens]{url}  % DO NOT CHANGE THIS
\usepackage{graphicx} % DO NOT CHANGE THIS
\usepackage{natbib}  % DO NOT CHANGE THIS AND DO NOT ADD ANY OPTIONS TO IT
\usepackage{appendix}
\usepackage{algorithmic}
\usepackage{amsmath,amssymb,amsfonts}
\usepackage{subcaption} 

\usepackage{xcolor} % Required for color definitions
\usepackage[
    final,
    colorlinks=true, % Activate colored links
    linkcolor=blue,  % Color for internal links
    citecolor=blue,  % Color for citations
    urlcolor=blue    % Color for URLs
]{hyperref}

\usepackage{caption} % DO NOT CHANGE THIS AND DO NOT ADD ANY OPTIONS TO IT
\usepackage{algorithm}
\usepackage{algorithmic}

\usepackage{newfloat}
\usepackage{listings}
\DeclareCaptionStyle{ruled}{labelfont=normalfont,labelsep=colon,strut=off} % DO NOT CHANGE THIS
\floatstyle{ruled}
\newfloat{listing}{tb}{lst}{}
\floatname{listing}{Listing}
 \nocopyright 
\title{Leveraging the Domain Adaptation of Retrieval
Augmented Generation Models for Question Answering and Reducing Hallucination}
\author {
    Salman Rakin\textsuperscript{\rm 1},
    Md. A.R. Shibly\textsuperscript{\rm 2},
    Zahin M. Hossain\textsuperscript{\rm 3},
    Zeeshan Khan,
    Dr. Md. Mostofa Akbar
}
\affiliations {
    salmankaderrakin@gmail.com, shibly.ar@gmail.com, zahinhasan2510@gmail.com, zeeshan@surroundapps.com, mostofa@cse.buet.ac.bd, 
}

\begin{document}

\maketitle

\begin{abstract}
While ongoing advancements in Large Language Models have demonstrated remarkable success across various NLP tasks, Retrieval Augmented Generation Model stands out to be highly effective on downstream applications like Question Answering. Recently, RAG-end2end model further optimized the architecture and achieved notable performance improvements on domain adaptation. However, the effectiveness of these RAG-based architectures remains relatively unexplored when fine-tuned on specialized domains such as customer service for building a reliable conversational AI system. Furthermore, a critical challenge persists in reducing the occurrence of hallucinations while maintaining high domain-specific accuracy. In this paper, we investigated the performance of diverse RAG and RAG-like architectures through domain adaptation and evaluated their ability to generate accurate and relevant response grounded in the contextual knowledge base. To facilitate the evaluation of the models, we constructed a novel dataset HotelConvQA, sourced from wide range of hotel-related conversations and fine-tuned all the models on our domain specific dataset. We also addressed a critical research gap on determining the impact of domain adaptation on reducing hallucinations across different RAG architectures, an aspect that was not properly measured in prior work. Our evaluation shows positive results in all metrics by employing domain adaptation, demonstrating strong performance on QA tasks and providing insights into their efficacy in reducing hallucinations. Our findings clearly indicate that domain adaptation not only enhances the models' performance on QA tasks but also significantly reduces hallucination across all evaluated RAG architectures. 
\end{abstract}

\section{Introduction}

Large Language Models (LLMs) generally store a vast amount of data encoded as factual knowledge in their parameters through fine-tuning on large corpora ~\cite{patrick_lewis_903a7121}. In recent years, LLMs have demonstrated significant advancements in natural language processing tasks including question answering, summarization, and dialogue systems. Besides, LLMs play a pivotal role in developing powerful agents, serving as essential components for reasoning and are fundamental for adaptation to new observations such as GPT in the context of Conversational AI \citep{qingyun_wu_0033ce6d}  \citep{jiarui_li_73b74c5a}. Recently, Retrieval Augmented Generation (RAG) models have demonstrated significant improvements compared to LLMs by leveraging external knowledge sources because of its capability to combine pretrained parametric memory with explicit non-parametric memory \citep{patrick_lewis_903a7121} \citep{xu2024search} \citep{johnson2023assessing}. Unlike the inherent built-in knowledge architectures of Large Language Models with limited access to external knowledge, RAG models have proven to be a promising approach for generating informative and context-aware responses. Recent advancements in the field of retrieval-augmented generation models have shown great potential for improving the factual accuracy and coherence of language models' responses rather than fine-tuning the entire LLM repeatedly \citep{rackauckas2024rag}.\\

Moreover, similar retrieval augmented architectures like REALM \citep{kelvin_guu_458fcab5} utilizes a fixed retriever component during the fine-tuning process and RETRO \citep{sebastian_borgeaud_e3604313} suggested that frozen BART embedding will be adequate for general-purpose retrieval from Wikipedia-like datasets. On the other hand, recent advancement on the original architecture of RAG, RAG-end2end \citep{shamane_siriwardhana_9061b788} observed that updating both the passage encoder and query encoder improves overall performance substantially in QA tasks across various domains like Covid-19, News etc. However, their works have been primarily focused on the role of domain-adapted retrievers without exploring any further impact of context awareness while achieving domain adaptation beyond general-purpose datasets.\\

RAG models possess the capability to generate more specific, diverse and factual language than parametric-only state-of-the-art architectures \citep{colin_raffel_92185fb9}. Although, RAG outperforms other parametric seq2seq models and task-specific retrieve-and-extract architectures, the performance of RAG models has been observed to be relatively limited when fine-tuned on specialized domains. Additionally, such architectures on existing research have  predominantly trained and evaluated on general-domain datasets, such as Wikipedia, which may not adequately capture the nuances and complexities of specialized domains. Hence, the effectiveness of RAG and RAG-based architectures has been largely under-explored in specialized domains such as customer service within the Hospitality Sector. \\

LLMs also suffer from a phenomenon widely known as "hallucination," where they generate content that is not grounded in the provided context or knowledge sources. Hallucinations in text-generation models have received significant attention recently \citep{joshua_maynez_66669e9a} while researchers have proposed various explanations for the occurrences of hallucinations in Large Language Models \citep{zhiying_zhu_5c6a6d1d} \citep{emily_m_bender_aa823910} \citep{collin_burns_a6054ee6}. These models have a tendency to generate outputs that are disconnected from the context documents, producing content that is plausible-sounding yet factually inaccurate. Recent efforts to address hallucination indicate that, combining parametric and non-parametric memory components tends to reduce hallucinations and enhance the interpretability of model outputs, particularly in knowledge-intensive tasks like question answering and text summarization \citep{jing_xu_5e014614} \citep{mojtaba_komeili_62a94f40} \citep{kelvin_guu_458fcab5} \citep{patrick_lewis_903a7121}. Shuster, Kurt, et al \citep{kurt_shuster_1c63c4b8} further addressed the issue of hallucinations within knowledge-grounded conversational tasks and illustrated that employing RAG architecture conditioned on retrieved documents reduced hallucination significantly. \\

Despite the potentials shown by Retrieval Augmented Generation models to address the hallucination problem by incorporating relevant external knowledge, the majority of the existing research didn't evaluate the impact of  domain adaptation on hallucinations. Besides, there have been limited understanding of their effectiveness in specialized, domain-specific applications such as Conversational AI for Customer Service. \citep{zijian_hei_8e084255}. In domains such as hotel auto-attendant services, generating misleading information can disrupt customer interactions and negatively impact the user experience, which ultimately diminishes trust in the service. Therefore, it would be valuable to further investigate the effectiveness of RAG architectures on a novel dataset composed of real-life hotel conversations and compare their performance between different state-of-the-art retrieval-augmented models.\\

Motivated by the recent advancements in the RAG architecture and to tackle the aforementioned research gaps, this paper undertakes three key initiatives: \textbf{(1)}  We explored the potential of leveraging domain adaptation techniques within a range of RAG and RAG-like  architectures and compared their effectiveness in generating responses that are both accurate and grounded to the context within the hotel-related conversational domains. \textbf{(2)} We further analyzed the impact of different RAG architectural variants on reducing hallucination through effective domain adaptation. Subsequently, we addressed the existing gap through an extensive categorized evaluation scheme on how domain adaptation techniques influence hallucination reduction. \textbf{(3)} We presented \textbf{\textit{HotelConvQA}}, a comprehensive dataset featuring domain-specific knowledge base comprising real-life hotel domain conversations and extended set of question-answer pairs based on those conversations.  This domain-specific data was then used for fine-tuning all the models to thoroughly evaluate the performance across all the RAG and RAG-like architectures in the hotel customer service domain, with a focus on improving response accuracy and mitigating hallucinations through effective domain adaptation.\\

The aim of our study is to investigate the potential of Retrieval Augmented Generation models on Question Answering Tasks in the context of domain-specific conversation dataset. Specifically, we focused on the hotel domain, as it represents a critical area for customer service where a reliable and accurate Question Answering System or Conversational AI can have a significant impact on user satisfaction and trust \citep{xiao_yang_45ea0d9b} \citep{shicheng_xu_c2104523} \citep{zijian_hei_8e084255} \citep{julien_delile_39a579c0}. Our findings highlight the impact of domain adaptation in enhancing the reliability and accuracy while building  conversational AI models and contributes to the field by providing open-source domain-specific datasets and detailed performance analyses, offering valuable insights into the practical application of RAG models in customer service. Our evaluation demonstrated strong results in all metrics across all the RAG variants including the Fusion in Decoder (FID) \citep{gautier_izacard_341465fa} model while fine-tuned on our hotel-domain dataset on QA tasks, providing  evidence into their efficacy in domain adaptation.  Finally, we further illustrated that an effective domain adaptation can potentially lead to reduce hallucination significantly across all RAG architectures. 

\section{Background and Related Work}\label{background}
Open-domain Question Answering (ODQA) represents a key challenge in the field of natural language processing and understanding \citep{kenton_lee_c414d8f8} \citep{patrick_lewis_52bc4f3e}. ODQA typically employs a two-stage architecture requiring contextual information to be fetched from adaptable knowledge base such as addressing complex datasets like WikiQA \citep{yi_yang_17e97143} and Natural Questions \citep{tom_kwiatkowski_493ab957}. Consequently, it utilizes a separate retriever component where relevant text segments are identified from the knowledge base based on input query followed by a reader or generator that processes the selected passages to generate final answers. In conventional setups, both the components are  optimized in isolation, with each model trained independently with question-answer (QA) pairs with respective ground truth passages as contexts. Historically, retrieval tasks have relied on sparse vector models like TF-IDF and BM25 \citep{stephen_robertson_66a90139}, but recent advancements have shifted towards dense text representations, such as Dense Passage Retrieval (DPR)\citep{vladimir_karpukhin_dd46c24e}, which utilizes dual BERT models \citep{jacob_devlin_c8d95bf8} to embed questions and passages based on semantic similarity. For answer generation, contemporary research has explored both extractive models, like BERT, and generative approaches, including BART\citep{mike_lewis_25a41328} and GPT-3 \citep{t__b__brown_7a189a92}. \\

ODQA system built on expensive large language models with billions of parameters such as GPT-3 \citep{t__b__brown_7a189a92} and Megatrone-LM \citep{deepak_narayanan_a27d4e8c} represents the complete knowledge as parametric memory. However, QA systems based on these large language models are prone to generating hallucinations, particularly when fine-tuned for downstream NLP tasks, where the models may produce inaccurate or fabricated information despite their extensive parameterization \citep{lei_huang_671937db}. Even with a two-stage pipeline that retrieves accurate information, the generator may still produce statements that are factually correct but misaligned with the contexts, often conflating similar entities or committing significant errors due to flawed token predictions \citep{jiarui_li_86997475} \citep{souvik_das_f2dd580b}. This issue of hallucination is a major hindrance to the safe deployment of these powerful language models in real-world applications that require reliable and truthful information \citep{s__m_towhidul_islam_tonmoy_5c21c3bb}. \\

Unlike ODQA with parametric only seq2seq model, RAG model can fine-tune its retrieval and generation components simultaneously, leading to strong performance on general QA tasks, outperforming both standalone extractive QA \citep{chen2017reading} and seq2seq neural language models \citep{patrick_lewis_903a7121}. Specifically, the RAG model has shown significant improvements in terms of answer accuracy and faithfulness as the retrieval component can provide relevant context to complement the language model's generation capabilities which potentially leads to reduce hallucinations. Although Previous works also suggested the effectiveness of RAG Architectures in conversational tasks \citep{kurt_shuster_1c63c4b8} and chatbot frameworks \citep{jing_xu_5e014614}, none of these works have explored the impact of domain adaptation on hallucination reduction. Subsequently, the RAG model has primarily been evaluated on open-domain datasets, such as those derived from Wikipedia, which limits its scope of applications like customer service.

\subsection{Retrieval Augmented Architecture}
Retrieval-Augmented Generation \citep{patrick_lewis_903a7121} models  have emerged as a promising solution to address the knowledge-intensive nature of NLP tasks \citep{xiao_yang_45ea0d9b} \citep{zijian_hei_8e084255}. Unlike the two-stage pipeline of traditional ODQA systems, Retrieval-Augmented Generation models utilized the neural-retrieval-in-the-loop technique with Dense Passage Retrieval \citep{vladimir_karpukhin_dd46c24e} and BART \citep{mike_lewis_8f70b723} sequence-to-sequence language model \citep{patrick_lewis_903a7121} \citep{kurt_shuster_1c63c4b8}. This method integrates a neural network architecture with transformers \citep{ashish_vaswani_5e53125f} as backbone consisting of an encoder-decoder framework.  In this system, the encoder processes the input query and the retriever converts it into a dense vector representation followed by employing maximum inner product search to identify and retrieve the most pertinent passages from an extensive corpus. These retrieved passages are then concatenated with the input query and fed into the decoder to generates final response. The entire system is typically optimized through end-to-end training, enhancing its capability to produce accurate and contextually relevant responses \citep{patrick_lewis_903a7121}. \\

Retrieval Augmented Architectures have gained significant attention for their scalable and interpretable design \citep{shamane_siriwardhana_9061b788}. Recent studies have significantly enhanced the capabilities of RAG models within dialogue-based systems \citep{rackauckas2024rag} \citep{bongsu_kang_41b0ebba}. It uniquely integrates retrieval and generation and merged in a single architecture, leveraging both a pre-trained BART generator and dense vector representations of Wikipedia articles indexed with the FAISS ~\citep{johnson2019billion} library capable of scaling to knowledge sources containing millions of documents. However, existing research on RAG models has focused only on open-domain tasks with Wikipedia-based datasets and there is limited understanding of their effectiveness in domain-specific applications, particularly in customer service area. Therefore, in this paper we aim to investigate the performance of different RAG and RAG-like architectures by leveraging domain adaptation techniques, with a focus on evaluating their ability to generate accurate and contextually appropriate responses in specialized domains.\\

\subsection{RAG-end2end Architecture}

The original RAG architecture maintains the encoding of passages and the external knowledge base in a fixed state, resulting in suboptimal performance when handling QA tasks that demand specialized knowledge beyond the scope of the original Wikipedia-based corpus \citep{tom_kwiatkowski_493ab957} \citep{joshi2017triviaqa}. To address the questionable performance of the RAG model on domain specific datasets, RAG-end2end \citep{shamane_siriwardhana_9061b788} proposes an extension to the RAG architecture that enables end-to-end training of the retriever and generator components. The key idea of RAG-end2end is to update the embeddings of the external knowledge base (documents) asynchronously during fine-tuning, in addition to jointly optimizing the parameters of the retriever and generator components. Furthermore, RAG-end2end introduces an auxiliary training signal that forces the model to reconstruct a given input sentence by retrieving relevant information from the external knowledge base \citep{shamane_siriwardhana_9061b788}. \\

Unlike RAG, RAG-end2end architecture iteratively re-encode and re-index the knowledge base during each training loop. The optimized architecture has been evaluated on datasets including CORD-19 \citep{wang2020cord}, Conversations \citep{yi_yang_17e97143} and News \citep{trischler2016newsqa}, demonstrating substantial improvements in accuracy by outperforming all other architectures. Although the conversation dataset, \textbf{QAConv}, comprises of business emails, panel discussions and work channels, it did not capture a single specialized domain adaptation capabilities crucial for building applications for robust customer service. Furthermore, aside from response accuracy they did not evaluate any impact of domain adaptation on hallucination reduction in their experiments. This paper aims to explore the potential of retrieval-augmented generation models to address the hallucination problem through domain adaptation.  By leveraging a comprehensive dataset specific to hotel interactions, we further evaluated and compared the performance of RAG-end2end with other retrieval-augmented architectures, such as RAG-original, RAG-fine-tuned, and the Fusing in Decoder (FID).

\subsection{Fusion in Decoder Architecture}

The Fusion-in-Decoder \citep{gautier_izacard_341465fa} model provides an alternative approach to the RAG architecture for leveraging external knowledge sources during generation. Unlike RAG, FiD employs a multi-head attention mechanism to fuse the hidden states with the encoded representations from the retriever. Additionally  rather than processing the documents separately, it encodes them independently and concatenates their representations. In contrast to RAG, which combines document probabilities with the encoder's output, FiD considers all retrieved documents together in the decoder, allowing the model to attend to multiple documents simultaneously. Moreover, unlike some methods that dynamically adjust document probabilities, FiD maintains a fixed retriever during training, which simplifies the model and training process.\\

This architectural difference enables FiD to demonstrate superior performance on several question-answering benchmarks by efficiently integrating multiple documents. But, their work only focused on the performance considering the number of retrieved passages within their proposed architecture. Furthermore, they did not assess whether incorporation of external knowledge with language model outputs could potentially help to counter hallucinations by grounding the generation in factual information.
Consequently, In our study, we compared the domain adaptation performance of FiD with different RAG architectures. In this research we further examined and categorized the occurrences of hallucinations on our domain specific dataset.  Finally, our results demonstrated that in addition to improving response accuracy, domain adaptation enhances the mitigation of hallucination significantly, which is a critical aspect while building conversational AI systems.
\section{Data Collection and Domain Specific Dataset Setup}

To leverage the power of retrieval-augmented generation models for enhancing customer service in hotel domain,several data sources were considered in curating the target corpus, including publicly accessible customer service datasets in conversational format, internal corporate chat logs, online customer reviews, employee inquiries, and pertinent domain-specific literature. However, there aren't many available corpus that supports all the required functionalities and lacks diverse conversational aspects of typical hotel related activities. Existing similar datasets such as \textbf{HotelRec}\footnote{\url{https://github.com/Diego999/HotelRec}}, \textbf{TripAdvisor}, \textbf{515k-hotel-reviews-data-in-europe}\footnote{\url{https://www.kaggle.com/datasets/jiashenliu/515k-hotel-reviews-data-in-europe}}, primarily offer hotel reviews or recommendations but lack the structured, conversational format needed for question-answer pairs, which limits their applicability in tasks requiring dialogue or interactive retrieval mechanisms. 

\subsection{Data Collection and Preprocessing}\label{corpus description}
For this research, three key corpora were identified as primary data sources: \textbf{Multiwoz}\footnote{\url{https://huggingface.co/datasets/pfb30/multi_woz_v22}}, \textbf{TripAdvisor}\footnote{\url{https://huggingface.co/datasets/argilla/tripadvisor-hotel-reviews}}, and a Local Hotel conversational dataset, with careful consideration of privacy and confidentiality issues. Subsequently, this data collection process was followed by a preprocessing phase, during which the datasets were standardized in format and cleansed of irrelevant or noisy information, facilitating the assessment of the domain adaptability of various models. The Multiwoz corpus formed the core dataset, focusing on hotel reservations, bookings, and recommendations, while TripAdvisor reviews and feedback introduced more variability with the integration of user reviews, complaints and feedback.  Additionally, we collected real-world conversations i.e. reservations, feedback, reviews, and issue resolution from a Local Hotel Dataset. This inclusion not only improved the authenticity and reliability of our overall dataset but also ensured that the dataset covers a diverse range of topics, such as making reservations, providing feedback and reviews, and addressing issues during a hotel stay. Although obtaining these conversations was challenging due to privacy concerns, data from local hotels was secured without any personally identifiable information.\\

After obtaining the raw datasets from the three corpora, custom scripts were used to segment the conversations into concise passages of around 100-120 words. This preprocessing step generated a knowledge base with 50,139 passages, capturing user-agent interactions. Subsequently, GPT-3.5-turbo APIs were utilized to create two outputs from the knowledge base: question-answer pairs and reconstruction signals. To expedite the data preprocessing phase, we incorporated six annotators to ensure the quality and relevance of the dataset, while also maintaining diversity within the hotel domain for the QA pairs and reconstruction signals.Details on the setup of our domain specific data, preprocessing steps using python scripts along with manual annotations has been described in the Appendix~\ref{sec:data_preprocess} and Appendix~\ref{sec:data_annotation}.

\subsection{Knowledge Base Generation}\label{kb generation}
To maintain consistent formatting across the training data, we preprocessed each corpus into a standardized conversational format between customers and hotel assistant. The MultiWOZ corpus is structured as a series of "turns," annotated with intent, belief states, and dialog acts for both user and system utterances. The MultiWOZ corpus, comprising 10,437 multi-turn dialogues, was restructured into 24,177 concise passages. Similarly, TripAdvisor hotel reviews were converted into a conversational format, producing 55,872 passages. To avoid bias toward a single corpus, the top 20,124 TripAdvisor passages were selected for inclusion. Additinally, for the 3,000 local hotel conversations, we removed any personally identifiable information and split them as mentioned before yielding 5,838 passages. Finally, we combined the MultiWOZ, TripAdvisor, and local hotel corpus to form our comprehensive Knowledge base, resulting in a total of 50,139 passages as described in table \ref{tab:data_distribution_corpora}.

\begin{table}[h!]
\centering
\resizebox{.85\linewidth}{!}{%
\begin{tabular}{ l r r r }
\hline
\textbf{Corpus}       & \textbf{KB}  & \textbf{QA}  & \textbf{Recon} \\ \hline
Multiwoz        & 24,177 & 8,678  & 7,283   \\ 
TripAdvisor     & 20,124 & 8,179  & 7,091   \\ 
Local Hotels    & 5,838  & 1,805  & 1,613   \\ 
\textbf{Total}  & \textbf{50,139} & \textbf{18,662} & \textbf{15,987} \\ \hline
\end{tabular}%
}
\caption{Knowledge Base, QA and reconstruction signal Distribution Across Corpora. Almost same number of passage, QA and reconstruction are taken from Multiwoz and TripAdvisor. Resourse from local hotels werer a bit scarce but still valuable. }
\label{tab:data_distribution_corpora}
\end{table}

\subsection{QA Pair Generation}\label{qa generation}
Our dataset includes a curated set of high-quality question-answer pairs. To generate these pairs, we utilized the GPT-3.5-turbo-0125 model through a custom script to produce comprehensive questions that encapsulate the key information from each passage in the knowledge base. We then applied the heuristic pruning Algorithm \ref{algo_pruning} to remove unnecessary or low-quality QA pairs, such as overly specific queries or generic responses like “Thank you” and “Goodbye”. This process resulted in a total of 18,662 QA pairs, which were sent to annotators for validation. Their primary task was to review each question to ensure alignment with the knowledge base and assess the overall quality. Once the validation phase was completed, we conducted an analysis of the entire dataset to examine the distribution of question and answer types. If any type appeared over-represented, for example if "what is the location of ....?" occurs more than necessary, we regenerated new questions that underwent the same validation process to maintain both quality and balance across the dataset. Through this process, we finalized THE the QA pairs and split with an 8:1:1 ratio for training all the RAG models for evaluation.

\begin{table}[h!]
\centering
\setlength{\tabcolsep}{25pt} % Adjust column separation (default is 6pt)
\begin{tabular}{ l r }
\hline
\textbf{Category} & \textbf{Count} \\ \hline
Train            & 14,929         \\ 
Test             & 1,867          \\ 
Validation       & 1,866          \\ 
\textbf{Total}   & \textbf{18,662} \\ \hline
\end{tabular}
\caption{The final set of QA pairs was split with an 8:1:1 ratio into training, testing and validation sets. }
\label{tab:dataset_distribution}
\end{table}

\subsection{Reconstruction Statement Generation}\label{recon generation}
The reconstruction signals, consisting of concise one-sentence summaries of longer passages from the knowledge base, are essential for improving the performance of the RAG end-to-end model as mentioned in the original paper. These signals were generated using the GPT-3.5-turbo-0125 language model for passages longer than 45 words, yielding a total of 15,987 summaries. Similar to the QA generation phase, these summaries were then forwarded to annotators for quality verification and to assess whether they captured the critical information contained in the original passages. Once validated, the reconstruction signals were combined with the QA pairs during the training phase of the RAG end-to-end model.

\subsection{Evaluation Dataset for Domain Adaptation and Hallucination}\label{eval generation}
To evaluate the domain adaptation capabilities and hallucination detection in the conversational agent, we curated two separate datasets: one for performance evaluation and another for hallucination detection. The performance evaluation dataset consists of ground truth QA pairs excluded from the training set, designed to assess the models on various question types described in table \ref{tab:question_distribution}. The difficulty levels were balanced with 62\% Easy, 30\% Hard, and 8\% Medium, ensuring a thorough evaluation across diverse formats and complexities. \\

\begin{table}[h!]
\centering
\begin{tabular}{ p{2.5cm}  r }
\hline
\textbf{Type}       & \textbf{\%}  \\ \hline
Extractive   & 62.00  \\ 
Abstractive  & 20.00  \\ 
Descriptive  & 14.00  \\ 
Comparison   & 2.00   \\ 
Calculative  & 2.00  \\ \hline
\end{tabular}
\caption{ Percentage of different types of questions present in our evaluation set.}
\label{tab:question_distribution}
\end{table}

For hallucination detection, We carefully constructed this dataset by injecting contradictory, false or misleading information into a subset of the conversations, creating scenarios where the model's responses would not align with the information stored in its parametric memory. We modified 63 passages and generated 90 QA pairs based on these altered passages. Additionally, we included \textbf{10} QA pairs that were not based on any passages in the knowledge base, to assess how the models would handle such out-of-domain scenarios. In total, the hallucination evaluation dataset comprised 100 QA pairs, allowing us to rigorously assess the hallucination-detecting capabilities of the RAG-like models.

\subsection{Dataset Analysis}\label{analysis}
The customer service dataset maintains a balanced distribution of questions, allows a re-validation indicator following manual quality assurance by the curators, ensuring the suitability and correctness of the training and evaluation data. By carefully examining the distribution of different question formats and corresponding answer types, we aimed to identify any potential biases or imbalances that could impact the performance of our models or misrepresent the underlying data. 
\subsection{Question Analysis}\label{question analysis}
\begin{figure}[h]
    \centering
    \includegraphics[width=0.5\textwidth]{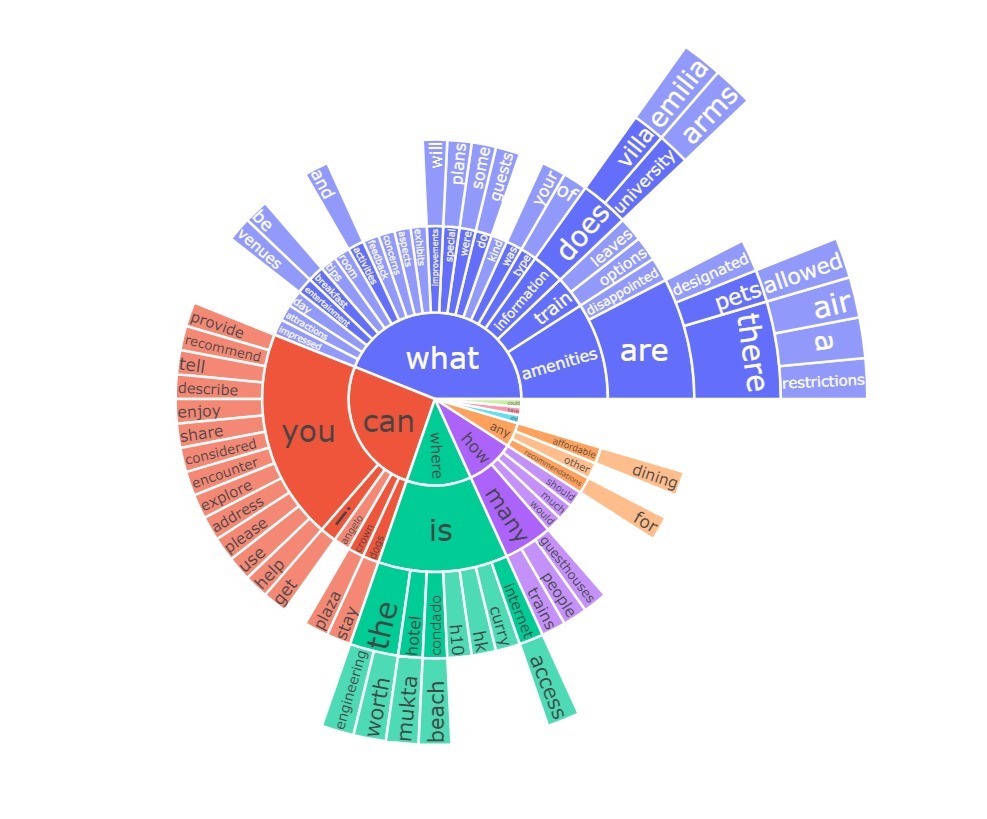} 
    \caption{Types of questions covered in \textit{HotelConvQA}. Types were determined by simply taking the first three words of a question. Comparison of which type occurs more are presented here. }
    \label{fig:question_distribution} 
\end{figure}

We performed an in-depth analysis of a sample of 270 QA pairs, uniformly selected from our knowledge base to ensure the presence of equal number of questions from all the corpora to assess the distribution of question types. This allowed us to categorize the questions into different types, such as “what”, “how”, “when” and so on.  By analyzing the frequency and distribution of various question types, we can better understand the breadth of the conversational scenarios. The distribution of these question types is visually represented in Figure \ref{fig:question_distribution}. 

\subsection{Answer Analysis}\label{answer analysis}
In our constructed evaluation set, the distribution of answer types reflects the varied nature of the questions posed to the models. The most prevalent answer types are information (24.07\%), recommendation (16.30\%), and experience (15.19\%), covering factual details about hotel amenities and policies, personalized suggestions tailored to user preferences, and experience-based answers that deliver subjective insights and impressions respectively. These categories are particularly significant as they represent key areas where nuanced and context-aware responses are critical, thereby serving as robust indicators of model performance.  Although less frequent, the remaining answer types in Table \ref{tab:answer_types} are also crucial for providing a well-rounded evaluation of model performance.\\

\begin{table}[h!]
\centering
\begin{tabular}{|l|r|}
\hline
\textbf{Answer type} & \textbf{\%} \\
\hline
Location & 5.19 \\
Information & 24.07 \\
Recommendation & 16.30 \\
Reservation & 1.48 \\
Amenities & 7.04 \\
Service & 3.70 \\
Yes/No & 7.78 \\
Feedback & 7.41 \\
Experience & 15.19 \\
Issue & 3.33 \\
Policy & 8.15 \\
Transport & 0.37 \\
Price & 0.74 \\
Food & 1.11 \\
\hline
\end{tabular}
\caption{Distribution of Answer Types in Percentage in \textit{HotelConvQA}.}
\label{tab:answer_types}
\end{table}

\section{Experimental Analysis}\label{exp}

\subsection{Model Architectures and Training Procedure}

We used the HuggingFace\footnote{\url{https://huggingface.co/}} \citep{alexander_m__rush_af39e9de}  library to implement all the RAG\footnote{\url{https://github.com/huggingface/transformers/tree/main/examples/research_projects/rag}} based architectures including the RAG-end2end\footnote{\url{https://github.com/huggingface/transformers/tree/main/examples/research_projects/rag-end2end-retriever}} model. To evaluate the effectiveness of all the models in the context of domain adaptation for the hotel domain, we conducted a series of experiments using our custom-built dataset. We utilized FAISS\footnote{\url{https://github.com/facebookresearch/faiss/wiki/Faiss-indexes}} \citep{jeff_johnson_00d12b47} for indexing and encoding our HotelConvQA knowledge base and opted HNSW FLAT as the indexing mechanism as instructed by \citeauthor{shamane_siriwardhana_9061b788} For the RAG-end2end model, we deployed eight NVIDIA Tesla V100 GPUs, each with 32 GB of memory, along with 96 vCPU cores on Amazon EC2 p3dn.24xlarge instances, facilitating parallel distribution of training, retrieval, and re-encoding tasks. This setup allowed for efficient workload distribution, with four GPUs dedicated to training and retrieval tasks, and the remaining four allocated for re-encoding and re-indexing tasks. Meanwhile, for other models— RAG-Finetuned, RAG-DPR, and Fusion-in-Decoder (FiD)—we maintained a consistent experimental setup using a single Tesla V100 GPU with 24 GB of memory as L4 GPU in colab, paired with an A100 GPU with 40 GB of memory and 432 Tensor Cores available in Google Colab. All the experiments were conducted on different hardware setups, chosen based on the specific needs of each model as. The experiments utilized distinct hardware setups and training parameters tailored to each model’s requirements, following the  recommendations of \citeauthor{shamane_siriwardhana_9061b788}. Additionally, the choice of training hyperparameters was considereed following the original RAG Paper by \citeauthor{patrick_lewis_903a7121} and further fine-tuned based on experimental validation results.\\

We initialized the DPR and BART compontens using two pre-trained open-source checkpoints from the HuggingFace model repository: rag-token-base\footnote{\url{https://huggingface.co/facebook/rag-token-base}} and rag-token-nq\footnote{\url{https://huggingface.co/facebook/rag-token-nq}}. Both models include a question encoder, retriever, and generator, with the rag-token-nq model pre-trained on the TriviaQA and NaturalQuestions datasets, enabling easier domain adaptation. The rag-token-nq model's retriever and generator were fine-tuned together on the WikiDPR QA dataset in an end-to-end manner, which generally allows it to generate more relevant responses. In contrast, the rag-token-base model, not fine-tuned on any downstream task, only computes predicted output loss for natural questions. However, we experimented with both of the checkpoints to compare and analyze response generation in the RAG-DPR and RAG-end2end settings. Below, we outline the training setup and methodology for each of the five models used in our experiments. \\

\textbf{RAG-Original :} The RAG-Original model served as our baseline for evaluation, following the architecture proposed by \citeauthor{patrick_lewis_903a7121} We indexed our HotelConvQA knowledge base and utilized this index to retrieve documents in response to queries. This model was evaluated without any kind of domain-specific fine-tuning, serving as a benchmark against which other models could be evaluated.\\

\textbf{RAG-Finetuned-QA :} The RAG-Finetuned-QA model builds upon the baseline RAG-Original architecture by incorporating domain-specific fine-tuning on our hotel domain dataset.  Like the RAG-Original model, it utilizes a Hotel Domain Knowledge Base, with dense vector representations indexed using the FAISS library. The loss function is designed to fine-tune both the generator and question encoder collectively. We have used ray as the distributed retriever with 4 retrieval workers as described in the original RAG paper. The model was trained for 20 epochs while the final checkpoint has been selected based on the highest validation accuracy. The purpose of this experiment was to assess the improvements gained from domain-specific training, comparing its performance against the baseline RAG-Original model.\\

\textbf{RAG-DPR-adapted :} The RAG-DPR-Adapted model investigates the impact of fine-tuning the Dense Passage Retriever (DPR) while keeping the pre-trained BART generator unchanged. This approach aims to enhance the retriever’s sensitivity to the domain-specific context of our hotel domain, allowing for more relevant passage retrieval. To understand the role of the retriever in the RAG architecture, we focused on fine-tuning only the retriever component. Since the DPR is pre-trained on Wikipedia-based datasets, we adapted it to the hotel domain using the Deep-Haystack\footnote{\url{https://github.com/deepset-ai/haystack}} framework. For each QA pair of the \textbf{HotelConvQA}, the corresponding passage was used as the positive context, along with 3 negative contexts, where the answer was not present, and 5 hard negative contexts, which were lexically similar but do not contain the answer. All these contexts were selected using the BM-25 lexical matching search as mentioned by the DPR author \citeauthor{vladimir_karpukhin_dd46c24e} The model was trained over 20 epochs with these contexts, resulting in two encoder checkpoints: the query encoder and the passage encoder. \\

Following this fine-tuning, we re-built the index using the passage encoder from the adapted DPR model. Then, we utilized the query encoder we got after finetuning the DPR to generate the embeddings of the questions. After that, we utilized the same approach of calculating dot product similarity score between those two embedding for getting relevant passaged based on the query and fed the top-$k$ retrieved passages to the generator.  After integrating the adapted retriever into the RAG model, we first evaluated its performance with the base model that had not been fine-tuned on any downstream tasks, and subsequently fine-tuned the merged model with QA pairs from our HotelConvQA dataset, noting results from both settings to assess improvements in retrieval and response generation. The consolidation of the fine-tuned DPR with the RAG architecture was intended to assess whether a domain-specific adaptation of the retriever, while preserving the general generation capabilities of the pre-trained BART model, could enhance performance on the hotel domain task.\\

\textbf{RAG-end2end :} The key idea behind RAG-end2end \citep{shamane_siriwardhana_9061b788} is to iteratively update the context encoder during training on the dataset of domain-specific question-answer pairs. Unlike the previous architectures, it updates the passage embeddings with the updated context encoder during each iteration of the training and then update the index of the external knowledge base asynchronously from the main training loop. Additionally, it requires reconstruction signals to inject more domain-specific knowledge into the model. This auxiliary signal was generated by summarizing the relevant passages which are factually correct and from the knowledge base and using them as a reconstruction target for the model. We used the training script provided by the RAG-end2end authors to fine-tune our model.  Finally, similar to RAG-DPR-adapted, to compare the impact of finetuning the entire setup with dowstream QA tasks, we initialized the rag-token-base, observing how the base model performs. After that, we initialized the model with rag-token-nq, which had already been fine-tuned on the question answering task. We trained the model for 10 epochs with a batch size of 16, mirroring the setup used in the previous experiments. \\

\textbf{Fusion-in-Decoder(FiD) :} The Fusion-in-Decoder~\citep{gautier_izacard_341465fa} model, Unlike the RAG models, which employ attention mechanisms to align retrieved passages with the input query, FiD adopts a simpler mechanism—concatenating the retrieved passages and processing them simultaneously in the decoder. In our fine-tuning process, we first generate embeddings for the knowledge base and then employ the same hotel domain QA dataset but in an alternative format. We utilize BM25 to retrieve 20 relevant passages including the gold passage and concatenated them as Context with each question-answer pair to build the final dataset for training. For the 19 relevant contexts other than the gold one, we re-ranked the passages using the BM25 scoring method and selected the top 19 based on this ranking. This approach allows us to follow the core methodology of the original FiD model, while tailoring the training process to our domain-specific needs. Following the approach outlined in the original Fusion-in-Decoder paper\footnote{\url{https://github.com/facebookresearch/FiD}}, which emphasized performance based on the number of retrieved passages, we concentrated on the end-to-end response accuracy, rather than separately evaluating retrieval performance.  The model was trained for 10 epochs using these retrieved contexts which took both the input question and its corresponding support passages, generating the final answer.\\
   
\subsection{Evaluation Setup}

\subsubsection{Accuracy Metrics} For evaluation, we used metrics such as Exact Match (EM), F1 score, and Top-k retrieval accuracy to assess model performance. The \textbf{EM} score measures the word-level match between predicted and actual answers while \textbf{F1} score calculates the overlap of words in the predicted answer with the reference regardless of order. For further evaluating the retrieval performance, we used \textbf{Top-k} retrieval accuracy, calculated by comparing the retrieved passages with the correct answer strings. We've measured top-5 and top-20 retrieval accuracy for evaluating retreival performance as followed by \citeauthor{shamane_siriwardhana_9061b788} in their experiments.\\

\subsubsection{Hallucination Evaluation Metrics} 

Existing benchmarks on hallucination evaluation primarily focus on assessing language models' abilities to either generate factual statements or distinguish them from non-factual ones. For example, \textbf{TruthfulQA}~\citep{stephanie_lin_db2fa70c} carefully designs to elicit imitative false statements with a high likelihood on the training distribution, while \textbf{FActScore}~\citep{sewon_min_cd363987} examines the factual accuracy of the generated text which manually annotates long-form model generations to identify atomic factual statements. Moreover, \textbf{HaluEval}~\citep{junyi_li_7d5da6be} and \textbf{FACTOR}~\citep{dor_muhlgay_07dc1398} involved hiring annotators to identify the presenece of hallucinations and then manually validated to ensure they meet certain requirements, such as being non-factual, fluent, and similar to factual completions. Recently \textbf{FactCHD}~\citep{xiang_chen_494751b9}, a dataset and benchmark identifies fact-conflicting hallucinations by leveraging evidence chains while a large-scale benchmark like HaluEval determines the hallucination propensity across various factuality patterns.\\ 

Despite the challenges of subjectivity across annotators and labor intensive annotation process, human evaluation continues to serve as the gold benchmark for reliably assessing hallucinations. \textbf{TruthfulQA}~\citep{stephanie_lin_db2fa70c} proposes a human-annotation guideline, which instructs evaluators to assign one of thirteen qualitative labels to the model output and verify answers by consulting a reliable source. Other approaches~\cite{nayeon_lee_5113f632}  human annotation to verify the validity of the proposed automatic evaluation metrics. \textbf{FActScore}~\citep{sewon_min_cd363987} requires annotators to assign three labels to each atomic fact: \textbf{``Supported''} or \textbf{``Not-supported''} for facts that are aligned by the knowledge source or not and \textbf{``Irrelevant''} for statements that are not related to the prompt. In our research, we adopted the \textbf{FActScore}~\citep{sewon_min_cd363987} approach, where our human annotators classified all responses from the five models into those  three categories. In addition to this, follwing previous works, the responses were further categorized into \textbf{Correct}, which refers to accurate and complete answers; \textbf{Ambiguous}\citep{tamkin2022task}, referring to unclear or open-to-interpretation responses; \textbf{Incompleteness}, where essential information was missing; \textbf{Bias}\citep{navigli2023biases}, indicating a preference or prejudice in the response; and \textbf{Under-informativeness}\citep{gao2023scaling}, which denotes vague or insufficient detail. The human evaluation results provide further insights into the performance of the models in reducing different types of hallucinations. A brief example of hallucinated responses from our hallucination evaluation settings for all the models has been described in Table \ref{tab:Model_response_example} in Appendix \ref{sec:hal_exm_details} .

\section{Results}\label{result}
We carefully constructed an evaluation set comprising human-generated QA pairs,designed to establish a neutral and unbiased setting, ensuring equal conditions by all means across the five model variants. The evaluation set contained both the ground truth answers for input questions for evaluation of the final response accuracy as well the title of the corresponding five passages including the gold passage to assess the retrieval performance. 

\begin{table*}[ht]
\begin{center}
\resizebox{.75\linewidth}{!}{%
\begin{tabular}{p{3.5cm}|c|c|c|c}
\hline
\textbf{Model} & \multicolumn{2}{|c|}{\textbf{Accuracy}} & \multicolumn{2}{|c}{\textbf{Precision (top-$k$)}} \\
\cline{2-5} 
\textbf{Name} & \textbf{\textit{EM}} & \textbf{\textit{F1}} & \textbf{\textit{Top-5}} & \textbf{\textit{Top-20}} \\
\hline
(1) RAG-Original & 4.00 & 10.92 & 11.20 & 14.50 \\
(2) RAG-Finetuned-QA & 12.50 & 30.29 & 20.00 & 23.70 \\
(3) RAG-DPR-adapted & 0.80 / 14.23$^{\mathrm{*}}$ & 6.41 / 31.54$^{\mathrm{*}}$ & 24.00 & 29.20 \\
(4) RAG-end2end & 1.40 / 17.36$^{\mathrm{*}}$ & 8.41 / 36.04$^{\mathrm{*}}$ & 24.40 & 29.20 \\
(5) Fusion-In-Decoder & 8.51 & 21.04 & - - & - - \\
\hline

% \multicolumn{5}{p{7cm}}{$^{\mathrm{a}}$Left column scores are based on models initialized with rag-token-base or not finetuned on downstream QA.} \\

\end{tabular}
}
\caption{Overall response accuracy and top-$k$ precision score of different RAG-based models after fine-tuning with domain specific dataset compared to RAG-Original, the baseline version of pre-trained RAG-sequence architecture on wikipedia datasets. The RAG-Original model demonstrates relatively low performance  whereas with the effect of domain adaptation the other models exhibit notable improvements in both response accuracy and retrieval performance across all evaluation metrics. For RAG-end2end and RAG-DPR-adapted models, left column scores are based on models initialized with rag-token-base and not being finetuned on downstream QA, whereas right column score represents models initialized with rag-token-nq and further finetuend on downstream QA tasks respectively.    }
\label{tab1}
\label{tab:domain_adaptation_results}
\end{center}
\end{table*}

\subsection{Overall Accuracy on Domain Adaptation}

The results presented in Table~\ref{tab:domain_adaptation_results} highlight the domain adaptation performance of the various Retrieval Augmented Generation based models on our hotel customer service dataset. The RAG-Original model serves as a baseline, achieving an EM score of \textbf{4.00} and an F1 score of \textbf{10.92}, reflecting the challenges of applying generic models to specialized domains without further adaptation. However, domain adaptation significantly boosts performance, as seen in the RAG-Finetuned-QA model, which shows a marked increase of \textbf{8.50} points in EM and \textbf{19.37} points in F1, underscoring the importance of domain-specific fine-tuning for improving response accuracy in specialized domains. The RAG-DPR-adapted model showed further improvements even over models directly fine-tuned on QA datasets, resulted in a more effective architecture for specialized tasks. The RAG-DPR-adapted model, integrating a domain-adapted Dense Passage Retriever and then initialized with BART, demonstrates notable improvements, achieving \textbf{1.73} points higher in EM and \textbf{1.25} points in F1 compared to the RAG-FinetuendQA model, particularly after fine-tuning for downstream QA tasks. However, without this fine-tuning, its performance dropped significantly, which further reflects the necessity of tailored QA finetuning for optimal results in specialized domains. Subsequently,
while the RAG-DPR-adapted model performs decently in terms of Top-5 (\textbf{24.0}) and Top-20 (\textbf{29.2}), it falls short in more stringent metrics like EM and F1, suggesting a challenge in generating
precise answers despite retrieving relevant information in the domain-adapted settings.\\
\begin{figure}
    \centering
    \includegraphics[width=0.9\linewidth]{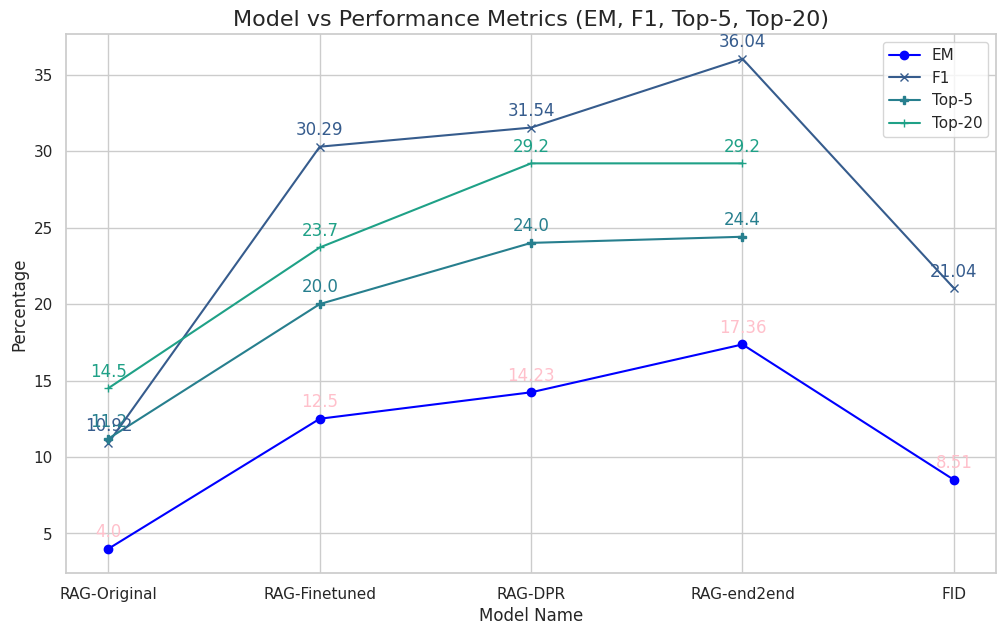}
    \caption{ The comparison graph showcases the improvements in retrieval accuracy across all models after being fine-tuned with the hotel domain dataset, as evidenced by notable gains in both Top-5 and Top-20 precision. It also indicates that increasing the number of retrieved documents leads to consistent improvements in retrieval performance which is evident by the parallel rise of Top-5 and Top-20 scores across all variants.}
    \label{fig:domain_adap_graph}
\end{figure}

As anticipated, The RAG-end2end model, which jointly fine-tunes both the retriever and generator with iterative updates to passage encoders, achieved the highest EM (\textbf{17.36}) and F1 (\textbf{36.04}) scores outperforming all models, indicates its ability to generate highly accurate and contextually relevant responses. However, similar to the RAG-DPR-adapted model, its performance significantly declined (\textbf{1.40} EM and \textbf{8.41} F1) when it has
been initialized with rag-token-base instead of rag-token-nq before training with our in-domain dataset, emphasizing the importance of appropriate initialization in specialized domains. Interestingly and very surprisingly, the Fusion-in-Decoder model, which has a more flexible architecture than the traditional RAG variants, does not demonstrate a clear advantage in this specific domain. As an additional benchmark, we included the Fusion-in-Decoder model, which integrates the retriever and generator components in a different architectural design. The FID model achieved an EM score of \textbf{8.51} and an F1 score of \textbf{21.04}, falling short of the performance of most of the fine-tuned RAG models except the RAG-Original.\\
\newline

\subsection{Retrieval Performance on Domain Adaptation}
\begin{table*}[ht]
\centering
\captionsetup{skip=5pt}
\resizebox{1.5\columnwidth}{!}{
\begin{tabular}{p{3cm}cccc}\hline
\textbf{Model Name} & \textbf{Supported} & \textbf{Non-Supported} & \textbf{Irrelevant}  & \textbf{Hallucination (\%)} \\\hline
(1) RAG-Original   & 71  & 29  & 0   & 29 \\
(2) RAG-Finetuned  & 77  & 21  & 2   & 23 \\
(3) RAG-DPR        & 80  & 18  & 2   & 20 \\
(4) RAG-end2end    & 85  & 10  & 5   & 15 \\
(5) FID            & 74  & 20  & 6   & 26 \\\hline
\end{tabular}
}
\caption{Hallucination Results on for the Models comparing the percentage of responses classified as supported, non-supported, and irrelevant for each model. RAG-Original has the highest hallucination rate, while the other models demonstrate improved performance in both supported responses and hallucination percentage. This comparison highlights the effectiveness of different model architectures and training strategies in reducing hallucinations and improving response accuracy.}
\label{tab:hal_results}
\end{table*}

The retrieval performance, measured by top-$k$ accuracy plays a cruicial role for assessing the models' effectiveness, as it reflects their ability to retrieve relevant information from a sizable knowledge base, which serves as the foundation for generating accurate as well as grounded responses to user inquiries.This metric captures the frequency with which the correct answer appears within the top 5 or top 20 passages, directly influencing the quality of generated responses. Figure~\ref{fig:domain_adap_graph} reveals that the RAG-Original model lags behind the other models in domain-specific tasks, with Top-5 and Top-20 retrieval accuracies of \textbf{11.20} and \textbf{14.50}, respectively. This lower retrieval performance underscores the limitations of the original architecture when applied to domain-specific tasks without prior fine-tuning or domain-specific adaptation. In contrast, both the RAG-Finetuned and RAG-DPR-adapted models exhibit substantial improvements in retrieval performance through finetuning, with the latter surpassing the RAG-Finetuned model by \textbf{4.00} and \textbf{5.50} points respectively.\\

Surprisingly, The RAG-end2end model, while excelling in generation metrics, shows limited gains in retrieval precision compared to RAG-Finetuned-QA and RAG-DPR-Adapted. This suggests that the domain-specific fine-tuning of the retrieval component, either individually or as part of an end-to-end optimization, can substantially enhance the models' ability to surface relevant information. The Fusion-in-Decoder model's performance is also shown, though not directly assessed for retrieval accuracy in this study. Although the retrieval capabilities of the FiD model have not been independently assessed in our experiments due to the unavailability of certain features in the original implementation of FID~\cite{gautier_izacard_341465fa}, the EM and F1 scores suggest that this model performs well by concatenating of retrieving relevant chunks of information to generate final response.

\subsection{Hallucination Reduction Performance}
Existing approaches \citep{stephanie_lin_db2fa70c,nayeon_lee_5113f632,sewon_min_cd363987} for evaluating hallucinations rely on human expert assessment as human evaluation ensures reliability. Therefore, we have employed human evaluation with the same annotators who contributed to dataset creation, ensuring consistency, credibility and precision. To enhance evaluation accuracy, we designed dedicated principles of agreement and re-validation described in Appendix \ref{sec:annotator_agreement} and developed specialized human evaluation guidelines presented in Appendix \ref{sec:annotator_guideline}, allowing annotators to  effectively evaluate each model-generated responses for all the different RAG architectures.  To ensure fair evaluation , we employed a blind annotation process, where the annotators are unaware of which model had generated each response, preventing any bias based on model identity. They worked independently under the predefined agreement, and in cases where over 50\% reached consensus on a classification, we accepted their judgment. For the opposite scenario, we conducted a manual review, reclassifying the outputs based on our observations.  Further details of entire Hallucination evaluation procedure and annotation mechanism along with the extensive classification of the inspecting categories are described in Appendix \ref{sec:hal_annotation}. \\

The results of our study reveal notable differences in how models handle supported and hallucinated responses, as well as in the quality and correctness of those supported answers. Table \ref{tab:hal_results} clearly demonstrates that domain adaptation reduces the rate for hallucination for all the models as the number of supported responses increased significantly after finetuning wiht our domain speciic dataset. However, it can be observed that, the reduction in hallucinations achieved with the cost of slight increase of irrelevant responses,  underscoring the inherent challenge of balancing hallucination mitigation with response precision. As anticipated, RAG-Original, with the highest hallucination rate of 29\%, exemplifies the model’s challenges in maintaining coherence between retrieved content and token prediction - a direct consequence of lacking domain-specific adaptation. In contrast, RAG-Finetuned exhibits a marked reduction in hallucinations\textbf{(6\%)}, signaling the efficacy of task-specific fine-tuning.\\

\begin{table}[ht]
\centering
\captionsetup{skip=5pt}
\resizebox{\columnwidth}{!}{
\begin{tabular}{lcccc}\hline
\textbf{Model Name} & \textbf{Not-Supported} & \textbf{Input} & \textbf{Fact}  & \textbf{Context} \\\hline
(1) RAG-Original   & 29  & 5  & 24   & 0 \\
(2) RAG-Finetuned  & 23  & 8  & 15   & 0 \\
(3) RAG-DPR        & 20  & 6  & 14   & 0 \\
(4) RAG-end2end    & 15  & 9  & 6   & 0 \\
(5) FID            & 26  & 1  & 14   & 0 \\\hline
\end{tabular}
}
\caption{Hallucination Results in Category of Occurrence. This table highlights how various models differ in their susceptibility to specific types of hallucinations, revealing areas for targeted improvement in model performance.}
\label{tab:hal_results_not_supported}
\end{table}

\begin{figure*}
  \centering
     \begin{subfigure}[b]{0.32\textwidth}
        \includegraphics[width=1\linewidth]{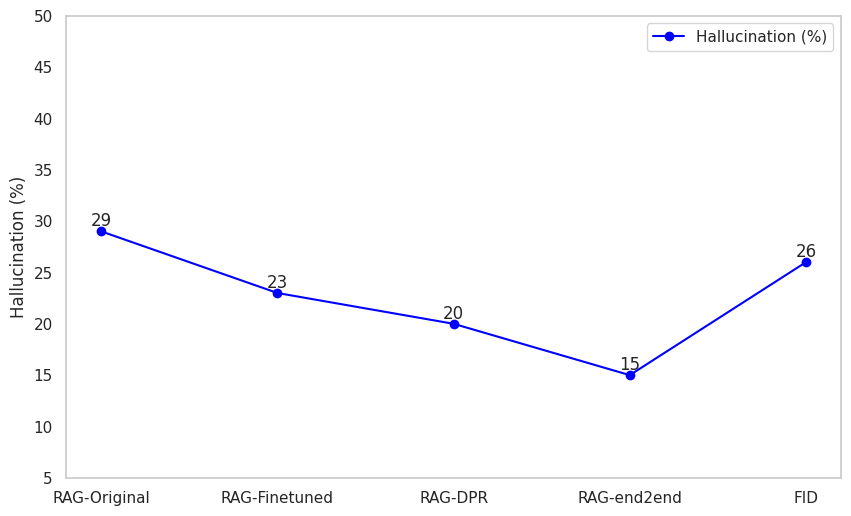}
    \caption{}
    \label{fig:hal_percentage}
    \end{subfigure}%
    \begin{subfigure}[b]{0.32\textwidth}
    \centering
    \includegraphics[width=1\linewidth]{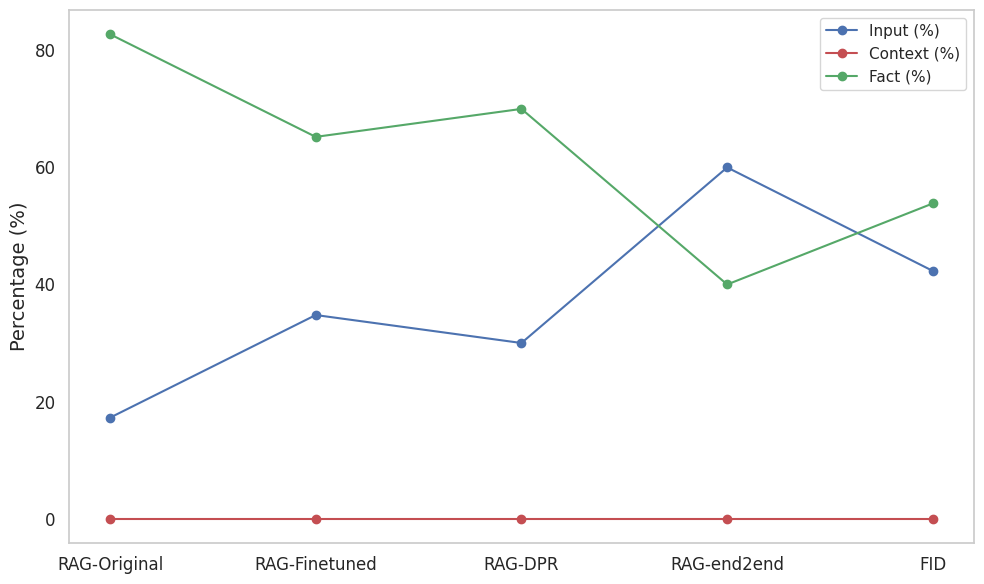}
    \caption{}
    \label{fig:hal_support_graph}
    \end{subfigure}%
    \begin{subfigure}[b]{0.32\textwidth}
    \centering
    \includegraphics[width=1\linewidth]{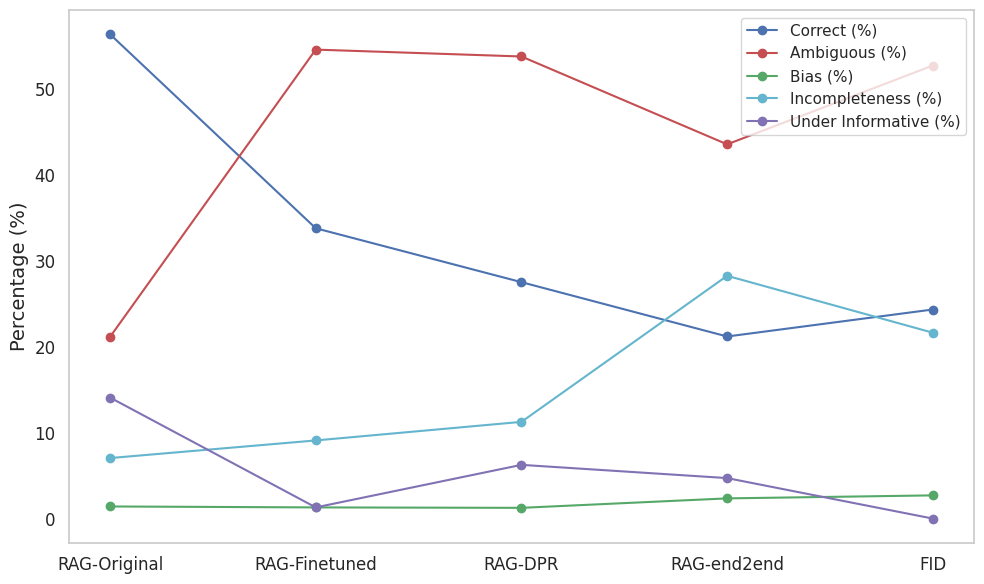}
    \caption{}
    \label{fig:hal_not_graph}
    \end{subfigure}%
    % \caption{Comparison of hallucination and non-hallucination categories across different models.}
    \captionsetup{justification=centering}
    \caption{From the left figure (a) represents the Hallucination Percentage Score of Models after Domain Adaptation. Figure (b) and (c) illustrates the comparison of Hallucination Percentage in Categories across models in Non-Supported and Supported Cases respectively.}
    \label{fig:hallucination_comarison}

\end{figure*}

Notably, the RAG-DPR model further refines retrieval accuracy by leveraging domain-adapted dense passage retrievers, reducing the hallucination rate even further compared to both RAG-Original\textbf{(11\%)} and RAG-Finetuned\textbf{(3\%)}, highlighted in the comparison graph \ref{fig:hallucination_comarison}\subref{fig:hal_percentage}. The lowest hallucination rate, 15\%, is observed in the RAG-end2end model, where both retrieval and generation components are fine-tuned simultaneously, again reflecting the benefits of holistic adaptation through re-encoding and re-indexing iteratively. Conversely, the FID model displays a relatively high hallucination rate of 26\%, alongside the highest occurrence of irrelevant responses, highlighting its limitations in handling domain-specific information accurately. \\

\begin{table*}[ht]
\centering
\captionsetup{skip=5pt}
\resizebox{1.6\columnwidth}{!}{
\begin{tabular}{lcccccc}\hline
\textbf{Model Name} & \textbf{Supported} & \textbf{Correct} & \textbf{Ambiguous}  & \textbf{Bias} &\textbf{Incomp.} &\textbf{Under-Inform} \\\hline
(1) RAG-Original   & 71 & 40 & 15 & 1 & 5 & 10 \\
(2) RAG-Finetuned  & 77 & 26 & 42 & 1 & 7 & 1 \\
(3) RAG-DPR        & 80 & 22 & 43 & 1 & 9 & 5 \\
(4) RAG-end2end    & 85 & 18 & 37 & 2 & 24 & 4 \\
(5) FID            & 74 & 18 & 39 & 2 & 16 & 0 \\\hline
\end{tabular}
}
\caption{Non-Hallucinated Results in Category of Occurrence. Breakdown of Supported responses of each model. This table presents a breakdown of non-hallucinated responses, classified into categories such as "Correct", "Ambiguous", "Bias", "Incomplete" and "Under-Informed". These results emphasize the varying quality of supported responses and highlight where models excel and where they struggle.}
\label{tab:hal_results_supported}
\end{table*}

A more granular analysis of hallucination type as shown in Table \ref{tab:hal_results_not_supported}, reveals that fact-conflicting responses dominate as majority of hallucinations fall under the category of Fact-Conflicting outnumbering those classified as Input Conflicting hallucination. Notably, the evaluators did not encounter any instance of context conflicting hallucination across all models, probably due to the ability of RAG Architectures to identify relevant contexts\citep{shi2023large} during the retrieval stage.  Another intriguing take away is that, apart from reducing the overall hallucination, domain adaptation substantially reduces fact-conflicting hallucinations but increases input-conflicting ones, a trend further visualized in Figure \ref{fig:hallucination_comarison}\subref{fig:hal_support_graph}. \\

We further classified the non-hallucinated responses to determine extended categories for the supported occurrences, summarized in the table \ref{tab:hal_results_supported}. Subsequently, a comprehensive comparative overview for all the models, emphasizing their performance across non-hallucination categories, highlighting their relative strengths and limitations for generating contextual response has been illustrated in Figure \ref{fig:hallucination_comarison}\subref{fig:hal_not_graph}. This comparison graph reveals interesting remarks that despite domain adaptation led to an increase in the number of supported responses, it also contributed to a consistent decline in the proportion of correct answers and a noticeable rise in ambiguous and incomplete responses. This figure also indicates a relatively stable bias rate across all models whereas the percentage of under-informative responses shows considerable fluctuation, suggesting a challenge of ensuring supported response while maintaining response quality.\\

Figure \ref{fig:hallucination_comarison}\subref{fig:hal_not_graph} further demonstrates that despite having lowest non-hallucinated response, RAG-Original model exhibits the highest correctness rate (\textbf{56.34\%}) but struggles with clarity (ambiguous) and detail(under-informative). Likewise,RAG-Finetuned demonstrated highest ambiguity rate (\textbf{54.55\%}) while RAG-DPR balances the correctness with lower level of ambiguity and incompleteness. The RAG-end2end demonstrates low correctness and high incompleteness, although it maintains a relatively lower level of ambiguity. In contrast, the FID model achieves a marginally better correctness rate (\textbf{24.32\%}) but struggles with higher ambiguity. An intriguing finding from Figure \ref{fig:hallucination_comarison}\subref{fig:hal_not_graph} suggests that despite a gradual increase in the percentage of supported responses, the correctness percentage drops consistently even after fine-tuning to different variants. This suggests that the domain-adaptation process, while improving the overall supported responses, may come at the cost of reduced factual accuracy and correctness in the generated outputs. This trade-off between support and correctness highlights the need for further refinement of the domain adaptation techniques to strike a better balance between these two important metrics.

\section{Discussion}

\subsection{Role of Domain Adaptation in RAG Architectures}
Our results highlight the impact of domain adaptation on enhancing their performance in specialized domains. The significant improvements observed in the RAG-Finetuned, RAG-DPR-adapted and RAG-end2end models, in terms of improved factual accuracy and reduced hallucination underscores the potential of these techniques to address the limitations of generic RAG models. The results clearly demonstrate that models fine-tuned through domain adaptation exhibit superior performance in generating contextually relevant and accurate responses compared to those that rely solely on general-domain training. The fine-tuning process also enabled the models to align more closely with the specific linguistic patterns, terminology, and contextual nuances of the hotel domain.  For instance, the domain-adapted models effectively incorporated the specificities of hotel operations, such as revenue figures and refurbishment details, into their responses, ensuring the significance of domain adaptation by generating content which is both accurate and contextually appropriate.

% \newpage
% \vspace*{150pt}
\subsection{Impact of reduced Hallucination with Domain Adaptation} 
Our experimental results in Figure \ref{fig:hallucination_comarison} clearly indicate that domain adaptation tend to reduce hallucination for all RAG architectures but at the expense of producing more ambiguous, incomplete and biased answers. This suggests that domain adaptation also introduced uncertainties in generating coherent, unambiguous, and complete responses, raising concerns about its effectiveness in high-stakes tasks like customer service where trust and reliability are crucial. However, the consistency in the production of almost bias-free and under-informative responses indicates that the models maintain a level of fairness and sufficiency despite other observed challenges. The success of domain adaptation in reducing hallucinations underscores its significance in the development of reliable dialogue systems. By tailoring the model's training process to the specific characteristics of the target domain, it is possible to achieve a more grounded and trustworthy AI output, minimizing the risk of generating misleading or incorrect information.

\subsection{Role of Retrieval in Hallucination Reduction}

The RAG-DPR-adapted model’s superior performance in reducing hallucinations highlights that the ability of RAG models to retrieve relevant information from a domain-specific knowledge base was a key factor in mitigating hallucinations. The results in Figure \ref{fig:hallucination_comarison} highlight that models with more sophisticated retrieval strategies, such as RAG-end2end and RAG-DPR-adapted, were more effective in producing accurate, fact-based responses compared to models with less integrated retrieval mechanisms, like RAG-Original and FID. In our experiments, we found that the models employing dynamic and context-aware retrieval processes were able to retrieve precise data points, such as hotel revenue figures or refurbishment costs, thereby reducing the likelihood of generating incorrect or speculative responses. Moreover, hallucinated responses for all the 10 the out-of-domain QA pairs suggest that the effectiveness of the retrieval mechanism is closely linked to the quality and relevance of the underlying knowledge base. When the retrieved contexts are comprehensive and well-aligned with the domain, it significantly contributed to reducing hallucinations by providing a robust foundation of facts for the model to draw upon. Conversely, when the context documents are less relevant or sparse, the models had to rely more on their internal parametric memory, which tends to produce more ungrounded content.

\subsection{RAG-end2end: Optimal Performance vs. Practical Limitations}
While RAG-end2end model significantly outperformed all other architectures, including the Fusion-in-Decoder (FID) model, across all metrics of retrieval and answer generation performance, it faces substantial challenges in computational complexity and practical applicability. Furthermore, despite demonstrating superior performance in reducing hallucinations and effectively adapting to the hotel domain its computational demands raise concerns about large-scale deployment. In our experiment, training the  on the hotel domain dataset, consisting of  50k passages as knowledge base and 29k QA pairs including reconstruction signals, required nearly \textbf{Eight} hours per epoch with batch size of 12. This was achieved using a high-performance machine, specifically the \textbf{Amazon EC2 p3dn.24xlarge} instance comprises of total 8 \textbf{Tesla V100 GPUs} with 32 GB memory each. This level of computational time is a significant bottleneck, particularly when the knowledge base and the dataset grow, ultimately making the model increasingly impractical for real-world applications. The high resource requirements raises concerns regarding the feasibility for large-scale deployment, especially in environments where computational resources are limited or where cost efficiency is a priority.
\subsection{Utilizing RAG Architectures for Conversational AI}
The insights gained from our research have important implications for the development and deployment of reliable conversational AI systems, particularly in the context of customer service applications. The improved performance and reduced hallucination rates demonstrated in Figure \ref{fig:hallucination_comarison}  highlight the potential of RAG Architectures to be utilized as a complete dialogue systems through enhancing the quality and reliability of customer service interactions. In a customer service setting, where users often seek accurate and contextually relevant information, the ability of RAG models to retrieve and generate responses that are tailored to the specific domain can lead to more satisfactory and trustworthy interactions. The reduction in hallucinations, especially in the Fact Conflicting category, can be particularly beneficial, as it helps to ensure that the information provided to customers is factually accurate and aligned with the organization's specific knowledge base. However, it is important to note that the utilization of RAG Architectures in Conversational AI for customer service settings also comes with its own set of challenges and considerations i.e. computational complexity, knowledge base maintenance and large-scale deployments in production.

\section{Conclusion and Future Work}  
In this paper, we have explored potential of domain-adapted retrieval-augmented generation (RAG) models in enhancing the performance of QA task. This research presents a comprehensive analysis by investigating the effectiveness domain adaptation and it's impact reduction of hallucinations in the RAG-like Architectures. Our study introduces several novel contributions to the field by presentint HotelConvQA, a detailed domain-specific dataset comprising a wide range of hotel-related conversations, which served as the foundation for evaluating the domain adaptation performance of various RAG and RAG-like architectures. Additionally, we conducted a detailed comparative analysis to understand their respective strengths and limitations in reducing hallucinations. While prior work has not explicitly determined the relationship between domain adaptation and hallucination reduction, our results emphasize that domain adaptation with adaptive retrieval mechanisms can significantly reduce hallucinations for all the RAG-based Architectures.\\[-1.1em]

Building on the insights gained from our paper, future research will aim to enrich the hotel domain dataset by incorporating a broader and more diverse array of scenarios, questions, and knowledge sources, facilitating more comprehensive evaluation of domain adaptation techniques. Subsequently, extending our analysis to include additional industry verticals such as healthcare, medicine etc would also provide valuable insights into the generalizability of the domain adaptation approaches explored in this work. Additionally, further research is needed by extending RAG as a component of conversational AI systems to evaluate their natural language understanding, dialogue management, and response generation capabilities.\\[-1.1em]

We have made all our efforts to implement and test the effectiveness of the RETRO~\citep{sebastian_borgeaud_e3604313} architecture on our hotel domain dataset using NVIDIA-Megatron and RETRO-pytorch frameworks; however, challenges related to system requirements and resource management impeded full integration. Therefore, future work could explore the comparison of RETRO with RAG-based models, enhancing our understanding of various retrieval-augmented strategies in handling domain-specific queries and reducing hallucinations. In the future, we aim to investigate the capabilities of RAG-based models in multi-hop question answering, extending their functionality beyond generating responses from a single document.\\[-1.1em]

We believe that our study will serve as a practical framework for leveraging retrieval-augmented generation models in high-stakes customer service applications, paving the way for more accurate and data-driven virtual assistants. Additionally, this research will be represented a significant step forward in the development of trustworthy and reliable conversational A systems, particularly for industry professionals seeking to manage and leverage dynamic business data effectively as the world changes rapidly. We hope that our contribution with the open-source hotel domain datasets and performance analyses presented in this study will establish a robust foundation for future research in this area, offering valuable discernment into the development efficient and effective conversational AI systems.

% \input{agreement}


\bibliography{aaai25}

\newpage

\appendix
\begin{appendices}

\twocolumn[
    \begin{center}
        \section*{Appendix: Domain Adaptation of RAG Models for QA and Hallucination Reduction}
        \vspace{1cm}
        
    \end{center}
]

\section{Data Preprocessing} \label{sec:data_preprocess}

\begin{figure}[h]
    \centering   
    \includegraphics[width=1\linewidth]{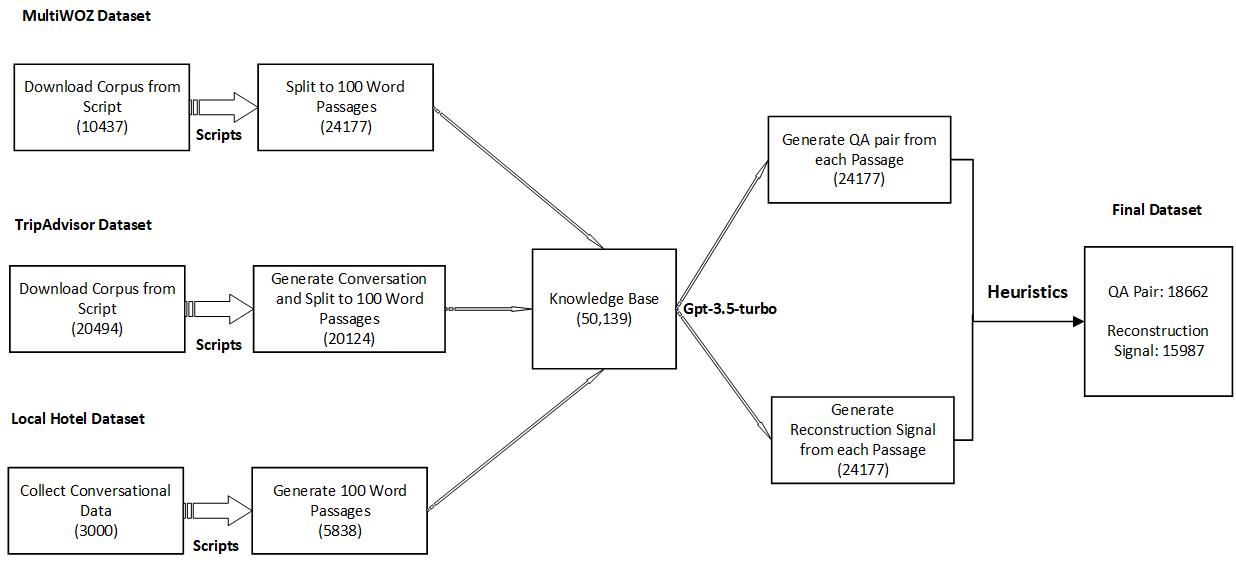}
    \caption{Workflow of Data Preprocessing Stages}
    \label{fig:data_pp_workflow}
\end{figure}

The data preprocessing phase involved consolidating three corpora to form a comprehensive knowledge base for the hotel domain. The MultiWOZ corpus contributed multi-domain conversations, including hotel bookings, which we segmented into 24,177 concise passages by processing it. Similarly, the TripAdvisor corpus enriched the knowledge base with detailed hotel reviews and descriptions, generating 20,124 passages. Before segmenting it into passages both corpora were converted into a conversational format using GPT-3.5-Turbo. Additionally, a local hotel dataset, derived from real-world customer interactions, provided 5,838 passages with domain-specific nuances essential for fine-tuning. 
All conversations were split into 100-120 word passages using custom scripts, resulting in a total of 50,139 passages in the unified knowledge base. We employed GPT-3.5-Turbo to generate both question-answer (QA) pairs and reconstruction signals from each passage.Throughout the entire preprocessing pipeline described in \ref{fig:data_pp_workflow}, They carefully reviewed and validated the QA paris, cross-checking them for accuracy and relevance to the hotel domain. This manual intervention was essential in maintaining the precision of the final dataset while preserving the conversational context vital for training our AI models. Annotators diligently refined the passages to ensure they accurately represented domain-specific interactions.

\section{Data Annotation Pipeline}
\label{sec:data_annotation}

\begin{figure*}
  \centering
     \begin{subfigure}{0.5\textwidth}
        \includegraphics[width=1\linewidth]{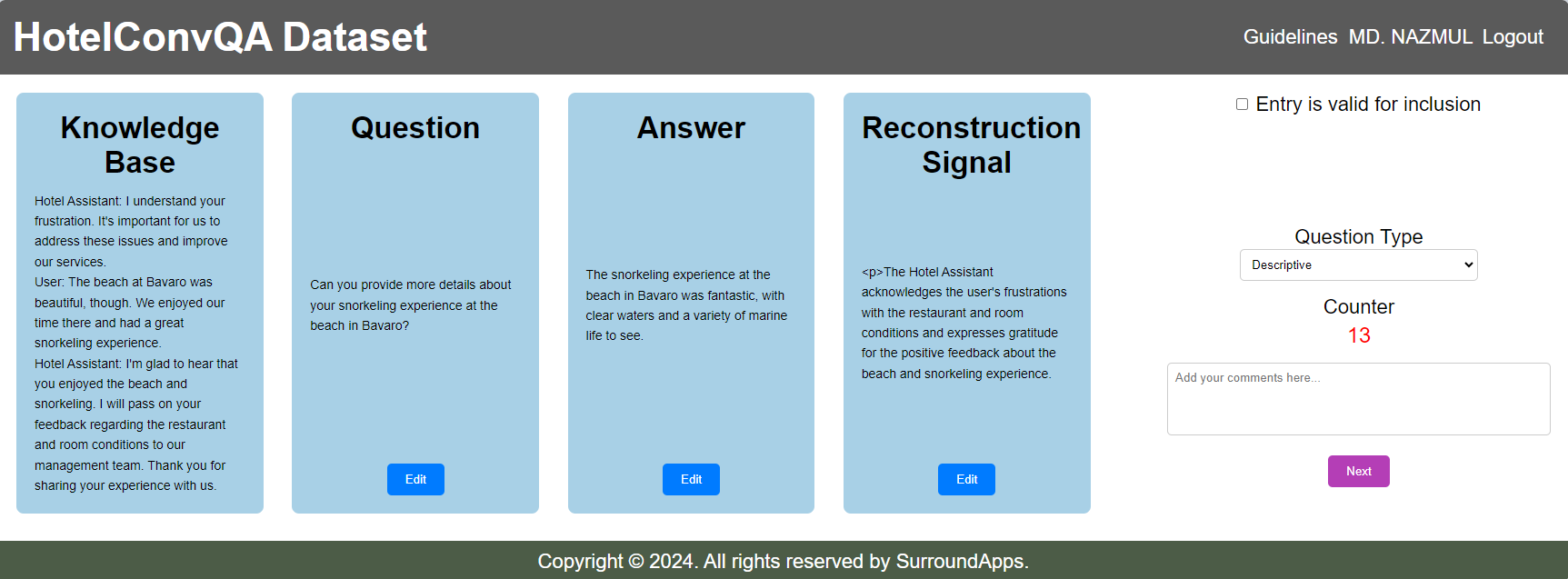}
    \caption{}
    \label{fig:app_data}
    \end{subfigure}%
    \begin{subfigure}{0.5\textwidth}
    \centering
    \includegraphics[width=0.85\linewidth]{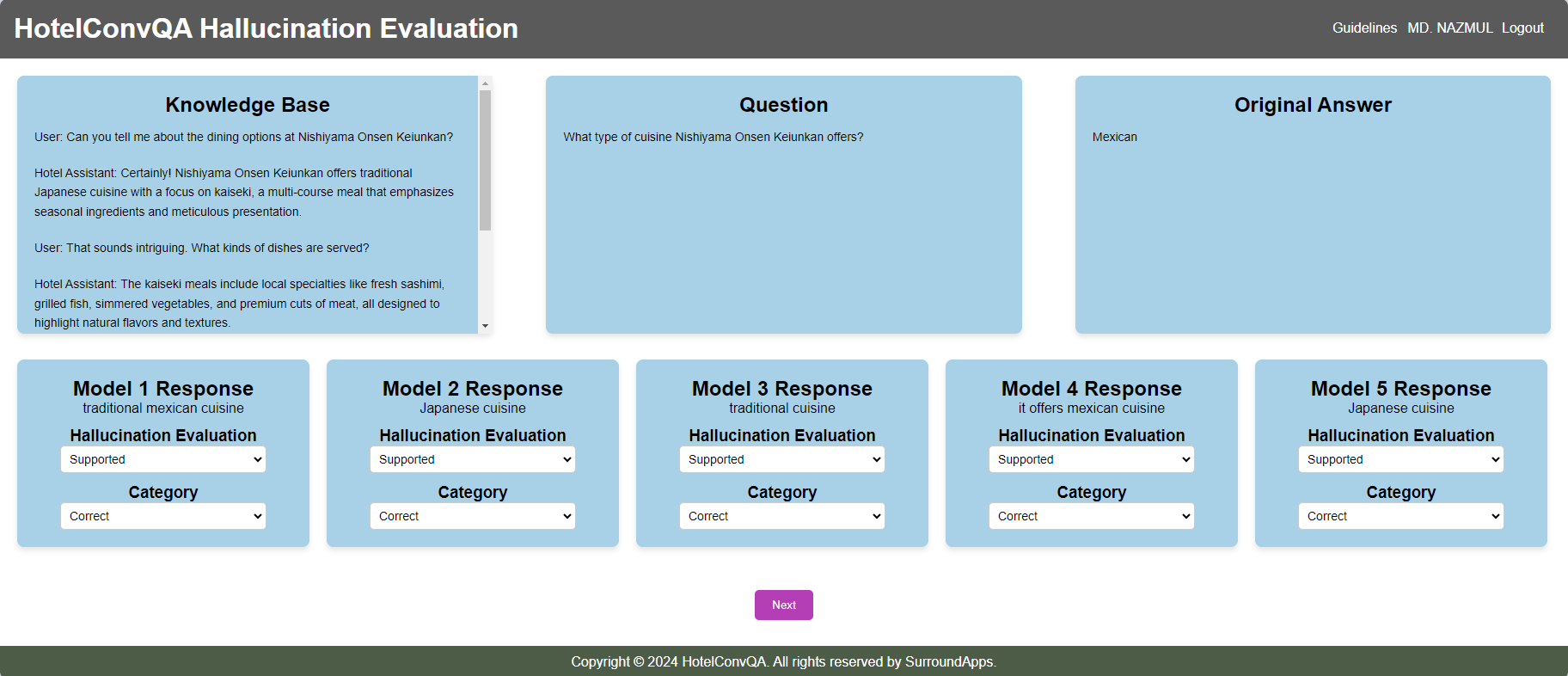}
    \caption{}
    \label{fig:app_hal}
    \end{subfigure}%
    % \caption{Comparison of hallucination and non-hallucination categories across different models.}
    \captionsetup{justification=centering}
    \caption{ Left: Screenshot of the annotator interface on QA Dataset Generation. Right: Annotator interface for hallucination evaluation and categorization. }
    \label{fig:application_interface}

\end{figure*}

To assist the manual annotators, we developed a data annotation pipeline and mechanism designed to expedite their assessment and quality assurance processes for the QA pairs and reconstruction signals, ensuring both efficiency and precision in evaluation.
In our data annotation process, we created an application providing a triplet, the passages with corresponding questions and answers illustrated in Figure \ref{fig:application_interface}\subref{fig:app_data}. Then we pruned the QA pairs using a hueristic algorithm \ref{algo_pruning}, reducing the QA pairs to 18,662 and reconstruction signals to 15,987. This ensured that only the most relevant and high-quality entries remained for model training. Then 

\begin{algorithm}[h]
  \caption{QA Pruning Algorithm}
  \label{algo_pruning}
  \begin{algorithmic}
    \REQUIRE $Q = \{q_1, q_2, \dots, q_n\}$: List of questions
    \STATE $A = \{a_1, a_2, \dots, a_n\}$: List of corresponding answers
    \STATE $K = \{k_1, k_2, \dots, k_m\}$: Set of keywords for pruning
    \STATE $P = \{p_1, p_2, \dots, p_p\}$: Set of phrases for pruning
    \STATE $L$: Minimum string length threshold
    \ENSURE Highlighted and pruned list of question-answer pairs
    \FOR{$i \gets 1$ to $n$}
        \IF{$\text{length}(q_i) < L$ \textbf{or} $\text{length}(a_i) < L$}
            \STATE Highlight $q_i$ and/or $a_i$ for length-based pruning
        \ENDIF
    \ENDFOR
    \FOR{each keyword $k \in K$}
        \FOR{$i \gets 1$ to $n$}
            \IF{$k \in q_i$}
                \STATE Highlight $q_i$ for keyword-based pruning
            \ENDIF
            \IF{$k \in a_i$}
                \STATE Highlight $a_i$ for keyword-based pruning
            \ENDIF
        \ENDFOR
    \ENDFOR
    \FOR{each phrase $p \in P$}
        \FOR{$i \gets 1$ to $n$}
            \IF{$p \in a_i$}
                \STATE Highlight $a_i$ for phrase-based pruning
            \ENDIF
        \ENDFOR
    \ENDFOR
    \FOR{$i \gets 1$ to $n$}
        \IF{$q_i$ \textbf{and} $a_i$ are both highlighted}
            \STATE Delete the pair $(q_i, a_i)$
        \ELSE[$q_i$ \textbf{or} $a_i$ is highlighted]
            \STATE Retain the pair with highlighting
        \ENDIF
    \ENDFOR
  \end{algorithmic}
\end{algorithm}

we recruited six annotators with an agreement described in Table \ref{tab:Agreeemt} for the task to refine and validate the generated 18,662 QA pairs with specific guidelines mentioned in Table \ref{tab:guideline_dataset} to ensure a more rigorous validation process. Their first task was to evaluate whether each question aligned with the knowledge base and whether it was of high quality. If the question met these criteria, it was retained. Otherwise, the annotators were instructed to rewrite the question.\\

% \begin{figure*}[h]
%     \centering   
%     \includegraphics[width=0.70\linewidth]{qa_eval.png}
%     % \caption{Screenshot of the annotator interface on QA Dataset Generation }
%     \label{fig:qa_val}
% \end{figure*}

% \begin{figure*}[h]
%     \centering
%     \includegraphics[width=0.70\linewidth]{hall_eval.png}
%     \caption{Screenshot of the annotator interface on QA Dataset Generation and Annotator interface for hallucination evaluation and categorization. }
%     \label{fig:example_halu}
% \end{figure*}

After manual evaluation and validation from the annotators we followed a series of re-validation steps outlined in Table \ref{tab:re-validation} on our end to maintain high quality of the final dataset of QA Paris and Reconstruction signals. Once all annotators reviewed the questions, we applied an agreement threshold: if over 50\% of the annotators agreed that the question was valid, we retained it in its original form. In cases where consensus was lacking, we selected the best version of the question from the rewritten suggestions. Following the validation phase, we analyzed the entire dataset to assess the distribution of question and answer types. If certain types were over-represented, we re-balanced the dataset by regenerating new questions of that type. These newly generated questions were again subjected to the same validation process by the annotators, ensuring both quality and balance across the dataset.

% \subsection{Data Annotation Guideline}\label{guidleine_dataset}

% \subsection{Re-validation steps by Dataset Analysis}

% \begin{figure}[H]
%     \centering   
%     \includegraphics[width=1\linewidth]{qa_eval.png}
%     \caption{Screenshot of the annotator interface on QA Dataset Generation }
%     \label{fig:qa_val}
% \end{figure}

\section{Hallucination Evaluation and Annotation Mechanism}\label{sec:hal_annotation}

After obtaining the model responses from the hallucination evaluation dataset, we sent the corresponding QA pairs containing the ground truth and model-generated answers to our annotators for further evaluation.  At each stage of the hallucination evaluation pipeline, the same team of six manual annotators carefully reviewed and validated question and answers, cross-checking them for coherence, domain relevance, and factual accuracy. We also provided a separate guideline described in Table \ref{tab:guidelin_hal} to them for properly evaluate the hallucinated responses from all the models on account of the same agreement with them before. The evaluation framework encompasses a multi-stage process, starting with the generation of responses, followed by a systematic analysis by human annotators. Figure  \ref{fig:application_interface}\subref{fig:app_hal} visually captures the flow of information from the initial input through to the final categorization of hallucinations, providing clarity on how outputs are scrutinized for inaccuracies.\\

% \begin{figure}
%     \centering
%     \includegraphics[width=1\linewidth]{hall_eval.png}
%     \caption{Application for evaluating hallucination by human annotators }
%     \label{fig:example_halu}
% \end{figure}

This process begins with the model generating answers based on external knowledge sources. After that the recorded responses from all the models undergo human assessment. Annotators are assigned with reviewing the content, comparing it to the ground truth and further determining whether hallucinations are present and then labelling every responses. The annotators were asked to perform a two-stage categorization as mentioned in the guideline. Initially, they categorized each response as either "Supported," "Non-supported," or "Irrelevant," based on its alignment with the knowledge base. Subsequently, they classified the responses into one of the following finer-grained categories: "Correct," "Bias," "Ambiguous," "Incompleteness," or "Under Informativeness. This categorization enables a granular analysis of where and how models deviate from the ground truth answers. Following this dual categorization, we applied a consensus rule: if more than 50\% of annotators assigned the same category, the classification was accepted. In cases where there was insufficient agreement, the responses were re-evaluated internally to ensure accurate categorization. This process was essential in assessing both the model’s performance and the types of hallucination it might generate. \\

\newpage
\section{Dataset and Evaluation Guidelines for the Annotators}
\label{sec:annotator_guideline}

% \vspace*{10pt}
\begin{table}[h]
\resizebox{\columnwidth}{!}{
\begin{tabular}{p{3cm}|p{9cm}}
\hline
\textbf{QA Dataset} &
\Large{%
\begin{itemize}
    \item Each annotator was tasked with validating and potentially rephrasing generated QA pairs based on the provided criteria.
    \item A streamlined pipeline was provided to assist annotators in their tasks. Each annotator received a passage along with GPT-generated question-answer pairs and a field for rephrasing the QA if necessary.
    \item The complete dataset was provided to all six annotators, ensuring that each contributed to the validation of a balanced selection of questions.
    \item Annotators were required to conduct quality checks on the generated QA pairs. If a pair was unclear or poorly phrased, they were responsible for rephrasing it to enhance clarity and fluency.
    \item If a generated QA pair did not align with the intended type (e.g., extractive, descriptive), annotators were encouraged to modify the pair to fit the appropriate category, ensuring alignment with the question-answer criteria.
    \item Annotators were instructed to avoid multi-hop questions, which require multiple passages. All questions were to be answerable based on a single passage.
\end{itemize}
}
\\ \hline
\end{tabular}
}
\caption{Guideline for the Dataset Generations for the Annotators}
\label{tab:guideline_dataset}

\end{table}

\begin{table}[h]
\resizebox{\columnwidth}{!}{
\begin{tabular}{p{3cm}|p{9cm}}
\hline
\textbf{Hallucination \newline Evaluation} &
\Large{%
\begin{itemize}
    \item Annotators were provided with the hallucination dataset, which included model responses. No model names were disclosed to ensure unbiased evaluation.
    \item Annotators were informed about the meaning of each category and required to categorize the questions in two stages:
    \begin{itemize}
        \item Firstly, questions were classified as "Supported," "Non-Supported," or "Irrelevant."
        \item Secondly, questions were evaluated for their correctness, with options including "Correct," "Ambiguous," "Incomplete," "Bias," and "Under-informativeness."
    \end{itemize}
    \item To enhance reliability, evaluations were cross-checked by ourselves to ensure consistency and accuracy in the categorization process.
\end{itemize}
}
\\ \hline
\end{tabular}
}
\caption{Guideline for Hallucination Evaluation}
\label{tab:guidelin_hal}
\hfill
\end{table}

\newpage
\section{ Agreement Criteria and Re-validation Steps}
\label{sec:annotator_agreement}

\begin{table}[h]

\resizebox{\columnwidth}{!}{
\begin{tabular}{p{3cm}|p{9cm}}
\hline
\textbf{Agreement} &
\Large{%
\begin{itemize}
    \item Annotators were compensated based on the number of question-answer pairs (QA) they processed and Hallucination Evaluation they classified into labels.
    \item A bonus system was implemented to reward high-quality contributions and for maintaining aspect ratio/diversity in the dataset.
    \item Annotators agreed to confidentiality and were prohibited from disclosing any details about the dataset.
    \item As long as quality was ensured, annotators may be required to rewrite QA pairs. They would receive payment for any rewrites performed.
    \item Adherence to the provided guidelines was mandatory for all annotators to ensure consistency and quality throughout the process.
\end{itemize}
}
\\ \hline
\end{tabular}
}
\caption{Agreement criteria with the Annotators in Evaluations}
\label{tab:Agreeemt}

\hfill
\\
\\

\resizebox{\columnwidth}{!}{
\begin{tabular}{p{3cm}|p{9cm}}
\hline
\textbf{Re-validation} &
\Large{%
\begin{itemize}
    \item Random sampling from each annotator was performed to check the quality. If the quality lagged at a certain level, tasks were reassigned.
    \item Repetitive questions were checked, especially when the number of repeated questions was not minor. 
    \item If there were many similar type of questions, such as yes/no answers, the dataset was sent back for rephrasing with newly generated questions.
    \item It was suspected that some annotators didn't properly read the passage. We found 4-5\% of the responses indicated misunderstanding of the passage, corresponding annotator was flagged for review.
\end{itemize}
}
\\ \hline
\end{tabular}
}
\caption{Re-validation Steps after Performing Manual Evaluation}
\label{tab:re-validation}

\end{table}

% Second row of tables - Hallucination Evaluation and Re-validation tables side by side

\newpage
\onecolumn
\section{Example of Response for Models}

\begin{table*}[h!]
\centering
\begin{tabular}
{|p{2cm}|p{2cm}|p{2cm}|p{2cm}|p{2cm}|p{2cm}|p{2cm}|p{2cm}| }
\hline
\textbf{Question} & \textbf{Answer} & \textbf{RAG original} & \textbf{RAG Fine-tuned} & \textbf{RAG DPR Fine-tuned} & \textbf{RAG-end2end} & \textbf{FiD} \\ \hline
Can you tell me about the kitchen amenities available at Outrigger Luana?	 & Stove, microwave, refrigerator provided. &	hotel assistant	& Outrigger Luana offers a variety of kitchen amenities such as a microwave, refrigerator, and coffee maker. &	Outrigger Luana offers a variety of kitchen amenities such as a microwave, refrigerator, and microwave for guests to enjoy. &	Outrigger Luana offers a variety of kitchen amenities such as a microwave, refrigerator, and coffee maker. & stove, microwave and refrigerator \\ \hline
What type of accommodation is the Alexander Bed \& Breakfast in the center of town? & 	The Alexander Bed \& Breakfast is a cheap, 4-star guesthouse.	& a guesthouse &	The Alexander Bed \& Breakfast is a 4-star guesthouse located in the center of town. &	The Alexander Bed \& Breakfast is a 4-star guesthouse with free parking and internet.	& The Alexander Bed \& Breakfast is a 4-star guesthouse located in the center of town. & the Alexander Bed \& Breakfast is 4-star guesthouse \\ \hline
In which floor of the Library Hotel does have a lounge area?	 & The second floor. & 	1st floor	& The second floor.& 	The second floor.& 	The second floor. &  second floor \\ \hline
What is the travel time from Bishops Stortford to Cambridge? & 	60 minutes.	 & 38 minutes & 	60 minutes. & 	60 minutes. & 	60 minutes. & 1 hour \\ \hline
\end{tabular}
\caption{ An Example of responses from all Models with Evaluation Set QA}
\label{tab:Model_response_example}
\end{table*}
\newpage
\section{Hallucination Example for Models} \label{sec:hal_exm_details}

\begin{table*}[h!]
\centering
\resizebox{18cm}{10cm}{
\begin{tabular}{|p{8.5cm}|p{2cm}|p{2cm}|p{2cm}|p{1.5cm}|p{1.5cm}|p{1.5cm}|p{2cm}| }
\hline
\textbf{Knowledge Base} & \textbf{Question} & \textbf{Answer} & \textbf{RAG original} & \textbf{RAG Fine-tuned} & \textbf{RAG DPR Fine-tuned} & \textbf{RAG-end2end} & \textbf{FiD} \\ \hline
User: Can you tell me about the dining options at Nishiyama Onsen Keiunkan? Assistant: Certainly! Nishiyama Onsen Keiunkan offers traditional Mexican cuisine with a focus on kaiseki, a multi-course meal that emphasizes seasonal ingredients and meticulous presentation. User: Mexican cuisine in Japan? That sounds intriguing. What kinds of dishes are served? Assistant: The kaiseki meals include local specialties like fresh sashimi, grilled fish, simmered vegetables, and premium cuts of meat, all designed to highlight natural flavors and textures. User: Where do they source their ingredients from? Assistant: Many ingredients are sourced from the surrounding mountains and rivers, ensuring fresh and authentic flavors. User: It sounds delightful! I can't wait to try it. & 	What type of cuisine Nishiyama Onsen Keiunkan offers?	& Mexican & traditional mexican cuisine		&  Japanese cuisine.	&  Traditional cuisine. &  it offers mexican cuisine	&  Japanese cuisine. \\ \hline
User: Hi, I recently stayed at your hotel of Marina Bay back in 2007 and had a great experience. I would definitely recommend it to others! Hotel Assistant: That's wonderful to hear! We always strive to provide a great experience for our guests. What did you enjoy most about your stay? User: I loved the fact that I was upgraded to a seaview room with a beautiful garden. The grounds and pool area were also well-maintained. Hotel Assistant: I'm glad you enjoyed your room and the amenities we offer. We take pride in keeping our property clean and welcoming for all guests. User: The staff at the hotel were very friendly and accommodating. Hotel Assistant: Thank you for your kind words. We noted that there was a reported issue with noise levels in the seaview rooms. Did that affect your stay? & How were the Wi-Fi connectivity issues managed in the MarinaBay Hotel room during the stay? & 	There were no Wi-Fi connectivity issues during the stay. The seaview room had good Wi-Fi signal and made the gues'ts convenient	&	managed &	The wi-fi connectivity issues in the Marinabay hotel room were resolved quickly and efficiently.	& The wi-fi in the room was not working properly, making it difficult to connect to the internet.	& There was no wifi issue during the stay.	&	The staff addressed the issues. \\ \hline
User: What can you say about the Dorchester in London? Hotel Assistant: The Dorchester, opened in 1940, is a prestigious luxury hotel in London. User: What amenities are available? Hotel Assistant: It offers elegant rooms, fine dining, and a renowned spa. User: What attractions are nearby? Hotel Assistant: It's located near Hyde Park and Mayfair. & When was Dorchester Hotel in London opened? & 	In 1940	&	1940 &	Early 1980s.	& Early 1980s. & 1940	&	The Dorchester Hotel in London opened in the early 1980s. \\ \hline
User: What's special about the Savoy Hotel in London? Hotel Assistant: The Savoy is known for its rich history and luxurious accommodations. User: What amenities does it provide? Hotel Assistant: It offers a renowned restaurant, a classic bar, and a theater. User: What attractions are nearby? Hotel Assistant: It's close to Covent Garden and the Thames River. It was the third Hotel in the Savoy group of hotels and restaurants owned by Royal family for over a century. &  Is Savoy Hotel is the first hotel in Savoy Group? & It's the third Hotel & is or isn't a hotel & Yes, it is. & Yes, it is. & no it isn't. & Savoy Group is the first hotel in Savoy Group. \\ \hline
\end{tabular}
}
\caption{Hallucination Example with QA pair }
\label{tab:hallucination_det_exmps}
\end{table*}

\end{appendices}

\end{document}